\documentclass[10pt,twocolumn,letterpaper]{article}

\usepackage{wacv}              

\usepackage{graphicx}
\usepackage{amsmath}
\usepackage{amssymb}
\usepackage{booktabs}
\usepackage{url}
\usepackage{booktabs}
\usepackage{adjustbox}
\usepackage{subcaption}
\usepackage{multirow}
\usepackage[accsupp]{axessibility}

\usepackage[pagebackref,breaklinks,colorlinks]{hyperref}

\usepackage[capitalize]{cleveref}
\crefname{section}{Sec.}{Secs.}
\Crefname{section}{Section}{Sections}
\Crefname{table}{Table}{Tables}
\crefname{table}{Tab.}{Tabs.}

\def\wacvPaperID{317} 
\def\confName{WACV}
\def\confYear{2024}

\begin{document}

\title{Assessing Neural Network Robustness via Adversarial Pivotal Tuning}
\author{
  Peter Ebert Christensen$^{1}$\quad 
  Vésteinn Snæbjarnarson$^{1}$\quad 
  Andrea Dittadi$^{2}$\\
  Serge Belongie$^{1}$\quad 
  Sagie Benaim$^{1}$ \\
$^{1}$University of Copenhagen \&
Pioneer Centre for AI
\qquad $^{2}$KTH Stockholm
}
\author{ Peter Ebert Christensen$^{1}$\quad 
  Vésteinn Snæbjarnarson$^{1}$\quad 
  Andrea Dittadi$^{2}$\\
  Serge Belongie$^{1}$\quad 
  Sagie Benaim$^{3}$ \\
$^{1}$University of Copenhagen
\qquad $^{2}$Helmholtz AI 
\qquad $^{3}$Hebrew University of Jerusalem
}
\maketitle


\begin{abstract}
The robustness of image classifiers is essential to their deployment in the real world. The ability to assess this resilience to manipulations or deviations from the training data is thus crucial. These modifications have traditionally consisted of minimal changes that still manage to fool classifiers, and modern approaches are increasingly robust to them. Semantic manipulations that modify elements of an image in meaningful ways have thus gained traction for this purpose. However, they have primarily been limited to \textit{style}, \textit{color}, or \textit{attribute} changes. While expressive, these manipulations do not make use of the full capabilities of a pretrained generative model. In this work, we aim to bridge this gap. We show how a pretrained image generator can be used to semantically manipulate images in a detailed, diverse, and photorealistic way while still preserving the class of the original image. Inspired by recent GAN-based image inversion methods, we propose a method called Adversarial Pivotal Tuning (APT). Given an image, APT first finds a pivot latent space input that reconstructs the image using a pretrained generator. It then adjusts the generator's weights to create small yet semantic manipulations in order to fool a pretrained classifier. APT preserves the full expressive editing capabilities of the generative model. We demonstrate that APT is capable of a wide range of class-preserving semantic image manipulations that fool a variety of pretrained classifiers. Finally, we show that classifiers that are robust to other benchmarks are not robust to APT manipulations and suggest a method to improve them.
\end{abstract}

\section{Introduction}

Significant work has been done in developing classifiers that work reliably in a broad range of data distributions~\cite{akhtar2021advances} with the aim of making them robust to corruption methods. A substantial part of this work considers robustness against adversarial $l_p$-bounded pixel-space perturbations. Since such perturbations act on raw pixels, they do not result in \emph{semantic manipulations} such as changes in lighting conditions or individual object textures.

Recently, a new generation of models has gotten much attention for their ability to generate highly expressive and photorealistic images. Models such as DALL$\cdot$E 2~\cite{dalle2}, RQ-VAE~\cite{lee2022autoregressive} and Stable-Diffusion~\cite{rombach2021highresolution} are capable of manipulating existing images with a high degree of detail and expressivity. In the context of neural network robustness, recent work has made use of these generative models to generate class-preserving semantic adversarial manipulations~\cite{song2018constructing, xu2020towards, poursaeed2021robustness, gowal2020achieving}, overcoming some of the limitations of the abovementioned pixel-space perturbations. However, these models are mainly constrained to specific styles or color changes. While the manipulations have proven important to assessing and improving robustness, they fall short of covering the entire space of possible class-preserving semantic manipulations. 

In this work, we aim to address this shortcoming by asking the following question: 
How can we leverage the full expressive power of a pretrained image generator to perform more general, highly detailed, photorealistic image manipulations for assessing the robustness of image classifiers? 
Given a pretrained classifier $C$ and a pretrained generator $G$, we wish to perform manipulations on a given set of images such that: (i) the resulting images are within the original dataset distribution, (ii) the manipulations are class-preserving, (iii) they fool the target classifier $C$, and (iv) they are highly expressive, i.e., the full capacity of the generator $G$ is leveraged.


\begin{figure*}
    \centering
    \begin{subfigure}[b]{0.015\textwidth}
    \adjustbox{varwidth=1cm,raise=.85cm}{{\rotatebox[origin=c]{90} {Input}}}
     \end{subfigure}
    \begin{subfigure}[b]{0.11\textwidth}
         \centering
         
         \includegraphics[width=\linewidth]{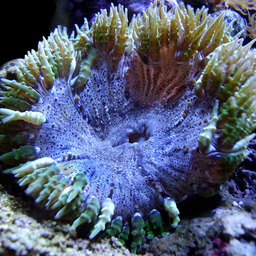}
     \end{subfigure}
     \begin{subfigure}[b]{0.11\textwidth}
         \centering
         \includegraphics[width=\linewidth]{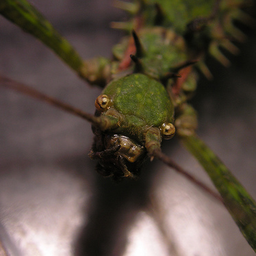}
     \end{subfigure}
     \begin{subfigure}[b]{0.11\textwidth}
         \centering
         \includegraphics[width=\linewidth]{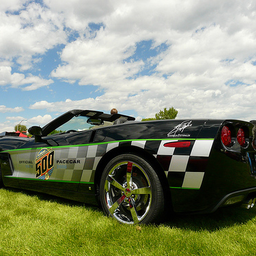}
     \end{subfigure} 
     \begin{subfigure}[b]{0.11\textwidth}
         \centering
         \includegraphics[width=\linewidth]{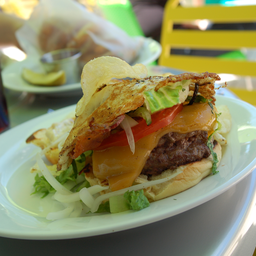}
     \end{subfigure} 
     \begin{subfigure}[b]{0.11\textwidth}
         \centering
         \includegraphics[width=\linewidth]{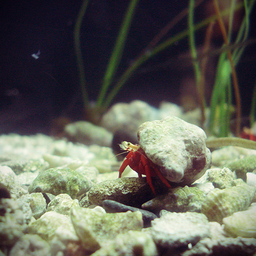}
     \end{subfigure} 
     \begin{subfigure}[b]{0.11\textwidth}
         \centering
         \includegraphics[width=\linewidth]{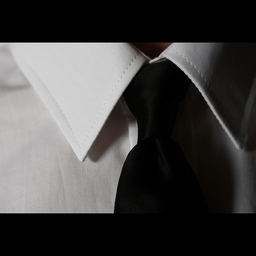}
     \end{subfigure} 
     \begin{subfigure}[b]{0.11\textwidth}
         \centering
         \includegraphics[width=\linewidth]{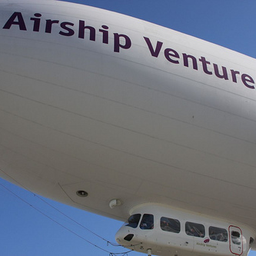}
     \end{subfigure} 
     \begin{subfigure}[b]{0.11\textwidth}
         \centering
         \includegraphics[width=\linewidth]{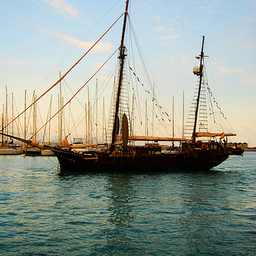} \\ 
     \end{subfigure} 
    
    \begin{subfigure}[b]{0.015\textwidth}
    \adjustbox{varwidth=1cm,raise=.85cm}{{\rotatebox[origin=c]{90} {Ours}}}
     \end{subfigure}
    \begin{subfigure}[b]{0.11\textwidth}
         \centering
         \includegraphics[width=\linewidth]{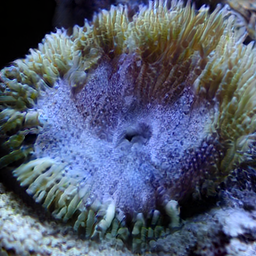}
     \end{subfigure}
     \begin{subfigure}[b]{0.11\textwidth}
         \centering
         \includegraphics[width=\linewidth]{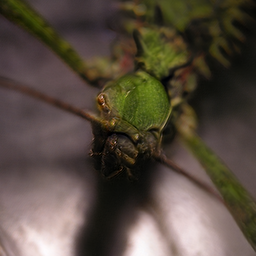}
     \end{subfigure}
     \begin{subfigure}[b]{0.11\textwidth}
         \centering
         \includegraphics[width=\linewidth]{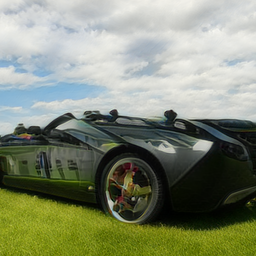}
     \end{subfigure} 
     \begin{subfigure}[b]{0.11\textwidth}
         \centering
         \includegraphics[width=\linewidth]{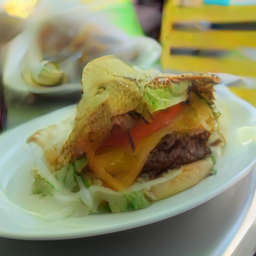}
     \end{subfigure}
     \begin{subfigure}[b]{0.11\textwidth}
         \centering
         \includegraphics[width=\linewidth]{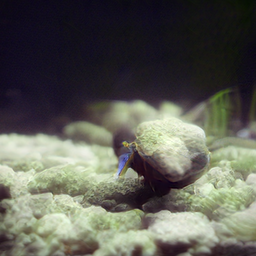}
     \end{subfigure}
     \begin{subfigure}[b]{0.11\textwidth}
         \centering
         \includegraphics[width=\linewidth]{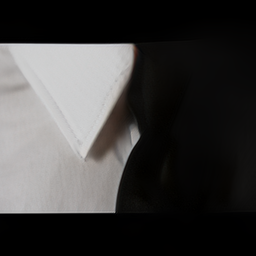}
     \end{subfigure}
     \begin{subfigure}[b]{0.11\textwidth}
         \centering
         \includegraphics[width=\linewidth]{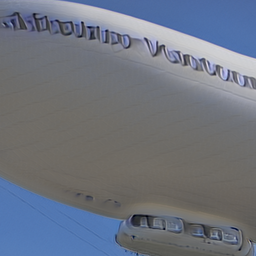}
     \end{subfigure}
     \begin{subfigure}[b]{0.11\textwidth}
         \centering
         \includegraphics[width=\linewidth]{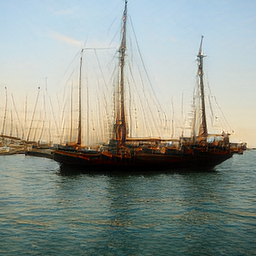}
     \end{subfigure}  
     
    \begin{subfigure}[b]{0.015\textwidth}
    \adjustbox{varwidth=1cm,raise=.85cm}{{\rotatebox[origin=c]{90} {\cite{lin2020dual}-pixel}}}
     \end{subfigure}
    \begin{subfigure}[b]{0.11\textwidth}
         \centering
         \includegraphics[width=\linewidth]{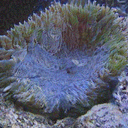}
     \end{subfigure}
     \begin{subfigure}[b]{0.11\textwidth}
         \centering
         \includegraphics[width=\linewidth]{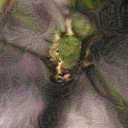}
     \end{subfigure}
     \begin{subfigure}[b]{0.11\textwidth}
         \centering
         \includegraphics[width=\linewidth]{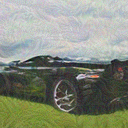}
     \end{subfigure} 
     \begin{subfigure}[b]{0.11\textwidth}
         \centering
         \includegraphics[width=\linewidth]{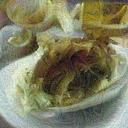}
     \end{subfigure}
     \begin{subfigure}[b]{0.11\textwidth}
         \centering
         \includegraphics[width=\linewidth]{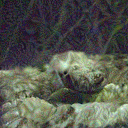}
     \end{subfigure}
     \begin{subfigure}[b]{0.11\textwidth}
         \centering
         \includegraphics[width=\linewidth]{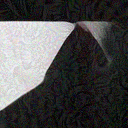}
     \end{subfigure}
     \begin{subfigure}[b]{0.11\textwidth}
         \centering
         \includegraphics[width=\linewidth]{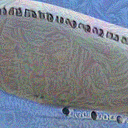}
     \end{subfigure}
     \begin{subfigure}[b]{0.11\textwidth}
         \centering
         \includegraphics[width=\linewidth]{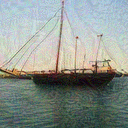}
     \end{subfigure}
     
\begin{subfigure}[b]{0.015\textwidth}
    \adjustbox{varwidth=1cm,raise=.85cm}{{\rotatebox[origin=c]{90} {\cite{lin2020dual}-latent}}}
    \end{subfigure}
    \begin{subfigure}[b]{0.11\textwidth}
         \centering
         \includegraphics[width=\linewidth]{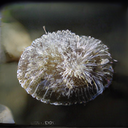}
     \end{subfigure}
     \begin{subfigure}[b]{0.11\textwidth}
         \centering
         \includegraphics[width=\linewidth]{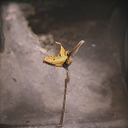}
     \end{subfigure}
     \begin{subfigure}[b]{0.11\textwidth}
         \centering
         \includegraphics[width=\linewidth]{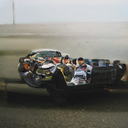}
     \end{subfigure} 
     \begin{subfigure}[b]{0.11\textwidth}
         \centering
         \includegraphics[width=\linewidth]{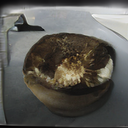}
     \end{subfigure}
     \begin{subfigure}[b]{0.11\textwidth}
         \centering
         \includegraphics[width=\linewidth]{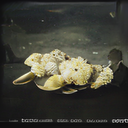}
     \end{subfigure}
     \begin{subfigure}[b]{0.11\textwidth}
         \centering
         \includegraphics[width=\linewidth]{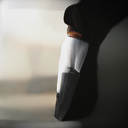}
     \end{subfigure}
     \begin{subfigure}[b]{0.11\textwidth}
         \centering
         \includegraphics[width=\linewidth]{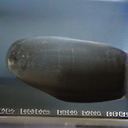}
     \end{subfigure}
     \begin{subfigure}[b]{0.11\textwidth}
         \centering
         \includegraphics[width=\linewidth]{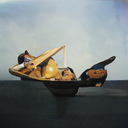}
     \end{subfigure}  

\begin{subfigure}[b]{0.015\textwidth}
    \adjustbox{varwidth=1cm,raise=.85cm}{{\rotatebox[origin=c]{90} {\cite{nulltextdiff}}}}
    \end{subfigure}
         \begin{subfigure}{0.11\linewidth}
         \centering
\includegraphics[width=\linewidth]{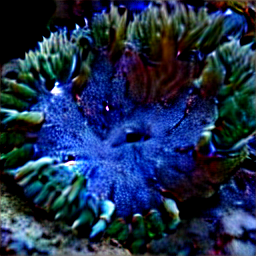}
     \end{subfigure}
     \begin{subfigure}{0.11\linewidth}
         \centering
         \includegraphics[width=\linewidth]{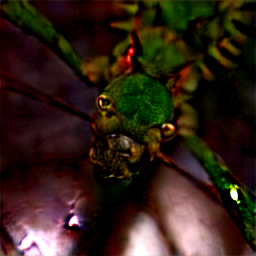}
     \end{subfigure}
     \begin{subfigure}{0.11\linewidth}
         \centering
         \includegraphics[width=\linewidth]{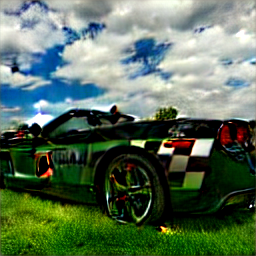}
     \end{subfigure} 
     \begin{subfigure}{0.11\linewidth}
         \centering
         \includegraphics[width=\linewidth]{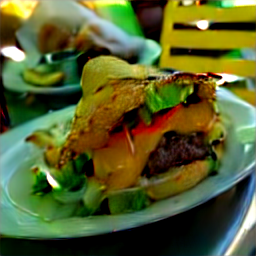}
     \end{subfigure}
     \begin{subfigure}{0.11\linewidth}
         \centering
         \includegraphics[width=\linewidth]{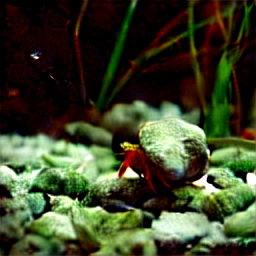}
     \end{subfigure}
     \begin{subfigure}{0.11\linewidth}
         \centering
         \includegraphics[width=\linewidth]{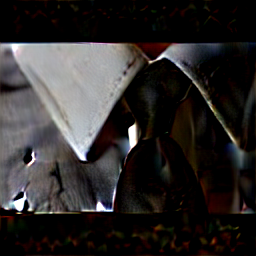}
     \end{subfigure}
     \begin{subfigure}{0.11\linewidth}
         \centering
         \includegraphics[width=\linewidth]{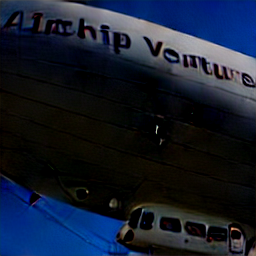}
     \end{subfigure}
     \begin{subfigure}{0.11\linewidth}
         \centering
         \includegraphics[width=\linewidth]{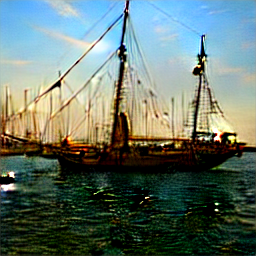}
     \end{subfigure}
     
\begin{subfigure}[b]{0.015\textwidth}
    \adjustbox{varwidth=1cm,raise=.85cm}{{\rotatebox[origin=c]{90} {\cite{song2018constructing}}}}
    \end{subfigure}
    \centering
    \begin{subfigure}{0.11\textwidth}
         \centering
\includegraphics[width=\linewidth]{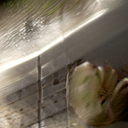}
     \end{subfigure}
     \begin{subfigure}{0.11\textwidth}
         \centering
         \includegraphics[width=\linewidth]{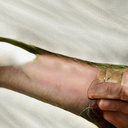}
     \end{subfigure}
     \begin{subfigure}{0.11\textwidth}
         \centering
         \includegraphics[width=\linewidth]{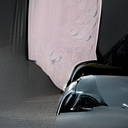}
     \end{subfigure} 
     \begin{subfigure}{0.11\textwidth}
         \centering
         \includegraphics[width=\linewidth]{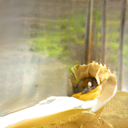}
     \end{subfigure}
     \begin{subfigure}{0.11\textwidth}
         \centering
         \includegraphics[width=\linewidth]{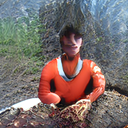}
     \end{subfigure}
     \begin{subfigure}{0.11\textwidth}
         \centering
         \includegraphics[width=\linewidth]{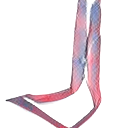}
     \end{subfigure}
     \begin{subfigure}{0.11\textwidth}
         \centering
         \includegraphics[width=\linewidth]{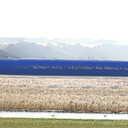}
     \end{subfigure}
     \begin{subfigure}{0.11\textwidth}
         \centering
         \includegraphics[width=\linewidth]{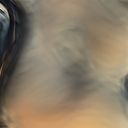}
     \end{subfigure}

    \caption{\textbf{Generated manipulations.} \textbf{Row 1} shows the original images.  \textbf{Row 2} shows our manipulations (using a distance of 0.2). \textbf{Row 3} shows the result of \cite{lin2020dual}, using pixel-space adversarial manipulations applied to StyleGAN-XL's reconstructions. \textbf{Row 4} shows the result of \cite{lin2020dual} with latent space manipulates applied using StyleGAN-XL.
    \textbf{Row 5} applies our method using a diffusion-based generative model instead of StyleGAN-XL. \textbf{Row 6} shows adversarially generated samples by \cite{song2018constructing} using StyleGAN-XL which are non-realistic and not class-preserving. 
    Our method manipulates images in a non-trivial but class-preserving manner, using the full capacity of the pretrained StyleGAN-XL generator. For example, it removes the eye of the mantis (second column), changes the type of race car (third column), changes the color of the crab tail (fifth column), removes the text on an airship (seventh column), and removes some of the ropes (eighth column). All of these are class-preserving examples that fool a pretrained PRIME-ResNet50~\cite{PRIME2021} classifier. In contrast, \cite{lin2020dual} either generates noisy and less realistic images (row 3) or images which differ significantly semantically and which do not preserve the input class (row 4).  
    }
    \label{fig:teaser}
    \vspace{-0.3cm}

\end{figure*}

We focus on the robustness of ImageNet classifiers and use the recently proposed StyleGAN-XL~\cite{styleganxl} generator, as it offers the ability to effectively manipulate style and content semantically. 
Our approach, \emph{Adversarial Pivotal Tuning (APT)}, 
first performs latent optimization to find the input pivot latent vector $w_p$ that results in the closest (but imperfect) reconstruction. We subsequently optimize the StyleGAN-XL weights with the following objectives: (1) reconstructing the image $x$, (2) while fooling the classifier $C$,
and (3) while ensuring the generated image appears real to the StyleGAN-XL discriminator, i.e., that it remains within the real image distribution.
To ensure that the manipulations are class-preserving, we bound the maximum perceptual distance between the input and generated image and stop the optimization when this distance is reached. 





Our method has the following main advantages:
(i) By generating images that are close to input images and appear realistic to a pretrained discriminator $D$, they are likely to be of high quality and fidelity as well as class-preserving. 
(ii) The manipulations are optimized over the entire space of the  generator parameters. 
By applying this optimization after an initial latent optimization stage, we ensure that the editing capabilities of StyleGAN-XL are preserved, thus allowing for expressive manipulations.  


We use the generated manipulations to assess the robustness of a variety of pretrained classifiers over a diverse range of architectures, as well as robust classifiers 
Our experiments show that there is a significant drop in classification accuracy of the image manipulated using APT and that these adversarial manipulations are transferable, indicating that the tested classifiers are not robust to them.
Visually (see \cref{fig:teaser}), we observe a wide variety of class-preserving image manipulations going beyond style transfer or changes in specific attributes. This is in contrast to \cite{lin2020dual}, which only utilizes StyleGAN-XL's latent space for manipulations. 
We subsequently consider an approach to improve the robustness to our manipulations through adversarial training on images that have been manipulated using our approach.

\section{Related Work}
\label{sec:rw}




\noindent\textbf{\noindent Semantic Adversarial Robustness.}
The majority of current literature considers adversarial robustness to pixel-space manipulations where the $l_p$ norm is bounded \cite{szegedy2013intriguing,fletcher2013practical}. For a comprehensive review, see \cite{akhtar2021advances}. We focus on an approach that semantically manipulates the input image, resulting in a naturally looking adversarial manipulation, instead of using noise attacks \cite{madry2017towards}. 

One set of works considers a specific class of semantic manipulations. These include geometric changes~\cite{xiao2018spatially,alaifari2018adef, engstrom2017rotation}, view changes~\cite{alcorn2018strike}, manipulating intermediate classifier features \cite{dunn2020evaluating,laidlaw2020perceptual,xu2020towards}, and inserting patches \cite{brown2017adversarial}. \cite{hendrycks2021natural} considers an image-filtering approach of natural images. 
Others consider the manipulation of style, texture or color statistics, where the image structure is fixed~\cite{hosseini2018semantic, bhattad2019unrestricted, shamsabadi2020colorfool}.  
Other works consider adversarial manipulation of facial attributes~\cite{joshi2019semantic, qiu2020semanticadv}. 
One can also consider deepfakes~\cite{tolosana2020deepfakes} as class-preserving semantic manipulations. 

Another line of work considers the use of pretrained generative models. \cite{song2018constructing} searches the latent space of a pretrained AC-GAN to find inputs that fool a given classifier. Unlike our method, they do not manipulate real images, resulting in less realistic generations and less faithful matching of the real image distribution---a result of AC-GAN's mode-dropping. \cite{xu2020towards} considers an autoencoder-based manipulation of real images, but it is restricted to style changes. 
\cite{gowal2020achieving} demonstrates an approach for adversarial training with samples generated by StyleGAN. However, it only manipulates a subset of the latent space variables, limiting the set of manipulations to coarse image changes. Moreover, our approach considers higher-resolution ImageNet samples while their approach is limited to low-resolution faces or MNIST digits. 
\cite{lin2020dual} projects images to a pretrained StyleGAN's latent space and adversarially manipulate their style code. Similarly, \cite{poursaeed2021robustness} manipulates both the style and noise latent vectors of StyleGAN. 
Our work takes a step further and manipulates not only the latent space of StyleGAN, but also its weights while preserving its editing capabilities. We thus enable the full utilization of StyleGAN's capacity to create highly expressive semantic manipulations, as shown in \cref{fig:teaser}. \\

\noindent\textbf{\noindent GAN Inversion and Image Manipulation.}
Our work is inspired by recent pretrained GAN inversion methods for manipulation of images. Some works optimize the latent space of a pretrained GAN~\cite{lipton2017precise, creswell2018inverting, abdal2019image2stylegan, karras2020analyzing} or use an encoder to find the latent input for a given image, such that the input image is effectively reconstructed~\cite{perarnau2016invertible, luo2017learning, guan2020collaborative}. 
In the context of StyleGAN, \cite{abdal2020image2stylegan++} showed that optimizing over StyleGAN's latent input space $\mathcal{W}$ results in unfaithful reconstructions. When considering optimization over the $\mathcal{W+}$ space, latent manipulations are inferior compared to the same manipulations over StyleGAN's $\mathcal{W}$ space. The W+ latent space is a more expressive space that consists of a concatenation of 18 $512$-dimensional $w$ vectors for each style of the AdaIn layer in the StyleGAN. 
To this end, \cite{roich2021pivotal} proposed to directly update StyleGAN's weights, following an initial latent optimization step. 
Unlike these methods, our goal is not to invert an input image, but rather to fool a classifier. 
\section{Adversarial Pivotal Tuning}
\label{framework}

\begin{figure*} [t] \centering
\includegraphics[scale=0.74]{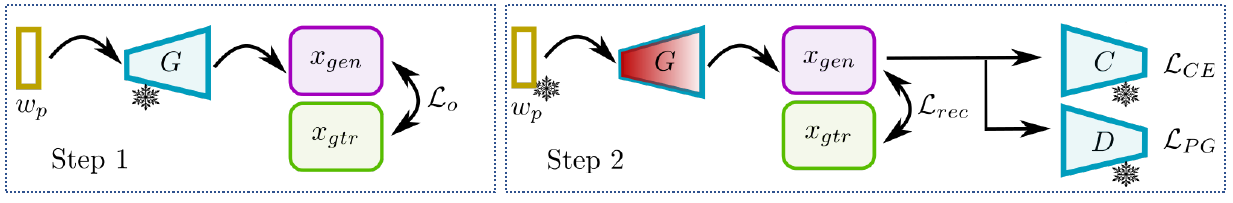}
\caption{\textbf{The Adversarial Pivotal Tuning (APT) framework.} In the first step, we optimize a style code $w_p$ using standard latent optimization $\mathcal{L}_o$ from \cref{eq:inversion}, while keeping the generator $G$ frozen. The loss is computed between the ground-truth image $x_{gtr}$ and the generated image $x_{gen}$. In the second step, we freeze $w_p$ and finetune $G$ (shown in red) using the three objectives from \cref{eq:loss}; a reconstruction objective  $\mathcal{L}_{rec}$, the projected GAN objective using the discriminator $D$, $\mathcal{L}_{PG}$, and our fooling objective $\mathcal{L}_{CE}$ using the classifier~$C$. A $*$ is used to indicate a frozen component.}
\label{fig:gan_fooling}
\vspace{-0.4cm}
\end{figure*}

Given a collection of images, we first filter those misclassified by $C$.
In order to fool $C$ on a correctly classified image, we wish to semantically manipulate it to be misclassified by $C$. 
Simple color jittering, rotation, translations, and semantically generated manipulations such as style, texture, or specific attribute change, can result in misclassification but remain limited in scope and realism.
We, therefore, suggest a new method, Adversarial Pivotal Tuning(APT), that learns non-trivial and highly non-linear image manipulations, while simultaneously ensuring the generated image stays within the data manifold. An overview of the APT method is shown in \cref{fig:gan_fooling}. Implementation details are provided in the supplementary material. 


We use StyleGAN-XL~\cite{styleganxl}, a generative model trained on ImageNet-1K, unless otherwise specified. The generator $G$ consists of a mapping network $G_m$ and a synthesis network $G_s$. The mapping network maps a random Gaussian latent variable $z \in \mathbb{R}^{64}$ along with a one-hot class label $c$ to the style code $w \in \mathbb{R}^{27 \times 512} = \mathcal{W}$.
The synthesis network subsequently maps $w$ and a noise vector $n$ 
to an RGB image $\hat{x} \in \mathbb{R}^{3\times H\times W}$ of height $H=256$ and width $W=256$. 
This generator is subsequently trained to fool a set of discriminators $\{D_l\}$ using the Projected GAN objective~\cite{Sauer2021NEURIPS}. We note that while StyleGAN-XL is used, our method is not limited to GAN based methods, and can be applied to other generative models such as diffusion-based methods as illustrated in Fig.~\ref{fig:teaser}. 


The first step of our method
aims at identifying a latent code $w$ (and noise vector $n$) that minimize the reconstruction error between a generated image $x_{gen}$ and a given ground-truth input image $x_{gtr}$, for a pretrained generator, in a similar manner to GAN inversion methods. This is done using the process of latent optimization over $w_p, n$:
\begin{align}\label{eq:inversion}
\underset{w, n}{\arg\min} \; \mathcal{L}_{\text {LPIPS }} \left ( x_{gtr}, G_s \left ( w, n; \theta \right ) \right )   &+ \lambda_n \mathcal{L}_n(n) 
\end{align}
Here, $x_{gen} = G_s(w, n; \theta)$ is the image produced by a \textit{pre-trained} synthesis network $G_s$ parameterized by weights $\theta$. We follow \cite{roich2021pivotal} and \cite{karras2020analyzing} in using a noise regularization term $\mathcal{L}_n$ and use $\lambda_n$ as a hyperparameter. The optimization is performed in  $\mathcal{W}$ space, as in \cite{roich2021pivotal}, and  
$\mathcal{L}_{\text{LPIPS}}$ is the perceptual distance introduced in \cite{zhang2018unreasonable}. 
In the second step, we modify the image $x$, to fool the classifier $C$, utilizing the full capacity of StyleGAN. We note that $G_s(w_p, n; \theta)$, i.e., the initial estimate for the reconstruction of $x$, should not be far from the adversarially manipulated image $\hat{y}$ we wish to generate. That is,  $w_p$ and $n$ are the results of the optimization of \cref{eq:inversion}.

We first consider the reconstruction objective, as in \cref{eq:pti_objective}, as we wish our manipulated image to be close to the input image $x$. Similarly to \cite{roich2021pivotal},
the generator weights are adjusted and regularized to restrict changes to a local region in the latent space, while the latent code $w_{p} \in \mathcal{W}$ and noise $n$ are fixed, leading to better reconstruction: 
\begin{align} \label{eq:pti_objective}
\hat{\theta} = \underset{\theta}{\arg\min} \; \mathcal{L}_{rec} ( x,  G_s  ( w_p, n; \theta ) ),
\end{align}
where $\hat{\theta}$ represents the new fine-tuned weights.
The reconstruction loss is defined as follows:
\begin{align}
\mathcal{L}_{rec} &= L_{pt} + \mathcal{L}_{R} \\ \mathcal{L}_{R} &= \mathcal{L}_{\text {LPIPS }}\left(x_r, x_r^{*}\right)+\lambda_{L 2}^R \mathcal{L}_{L 2}\left(x_r, x_r^{*}\right) \\
\mathcal{L}_{pt} &= \mathcal{L}_{\text {LPIPS }}\left(x, x_p^{*}\right)+\lambda_{L 2}^P \mathcal{L}_{L 2}\left(x, x_p^{*}\right) 
\end{align}
where $x_p^*$ is generated using the modified weights as $G_s(w_p, n; \hat{\theta})$.
A locality regularization term ($\mathcal{L}_{R}$) is applied by restricting changes to a local region in the latent space. Specifically, setting 
$w_r=w_p+\alpha \frac{w_z-w_p}{\left\|w_z-w_p\right\|_2}$, 
where, in each iteration, $z$ is sampled from a normal distribution and $w_z$ is obtained by applying the mapping network $G_m$ to $z$ and class $c$ of the input image $x$ and $\alpha$ is a hyperparameter.
We then generate $x_r$ and $x_r^*$ using the initial and modified weights, $G_s(w_r, n; \theta)$ and $G_s(w_r, n; \hat{\theta})$ respectively.


Secondly, we wish to fool the classifier $C$. That is, the cross-entropy loss for the classifier's prediction on the manipulated image should be high. In practice, we observed a more stable optimization when minimizing the cross entropy between the classifier's prediction and an incorrect class label chosen at random. 
Lastly, we utilize 
StyleGAN's pretrained discriminators $\{D_l\}$ to distinguish between real and synthetic images, and enforce that the manipulated image appears real.  Following  \cite{Sauer2021NEURIPS}, we consider the following objective: 
\begin{align}
    \mathcal{L}_{PG} = \sum_{l} \log \left(1-D_l\left(
    G_s(w_p, n; \theta\right)\right)) ,
\end{align}
where the weights of each $D_l$ are fixed. We note that as $\{D_l\}$ is not trained, and is used to assess generated images, one may use a different generator to StyleGAN  (for instance, a pretratined diffusion model).
We then finetune $G$'s weights $\theta$ with the following objective:
\begin{align}
    \mathcal{L}_{APT} &= \mathcal{L}_{rec} + \lambda_{CE} \mathcal{L}_{CE}(c_{any}, C(G_s(w_p, n; \theta))) \\ &+ \lambda_{PG} \mathcal{L}_{PG}
    \label{eq:loss}
\end{align}
where $c_{any}$ is a randomly chosen class different from the true class, and $\mathcal{L}_{CE}$ is the cross entropy loss. 
As both the classifier $C$ and discriminators $\{D_l\}$ are fixed, the generated image is changed to match the reference image as closely as possible, while deviating only slightly to change the class predicted by the classifier.
Given a desired maximum distance $d$, we consider generated images for which $\mathcal{L}_{pt} \leq d$, where $d$ is a hyperparameter, and stop the optimization whenever $\mathcal{L}_{pt} \geq d$. We note that, unlike traditional frameworks that use a maximum $l_p$ norm to bound adversarial examples, we consider $\mathcal{L}_{pt}$ which uses both a pixel-based distance and a perceptual distance.

\noindent\textbf{Computational Time\quad}
Using APT takes around 5 minutes and is bottlenecked by the GAN synthesis speed (but can be performed in parallel).  
While noise-based attacks are faster, we consider a different attack vector. 
\section{Experiments}
\label{sec:experiments}

We begin by assessing the degree to which our generated images 
(i) are within the ImageNet distribution, 
(ii) represent the same class as the corresponding input images (i.e., the manipulation is class-preserving),
(iii) fool a target classifier, i.e., the classifier misclassifies the generated images,
and (iv) exhibit a wide variety of semantic changes.

To this end, we consider a collection of pretrained classifiers including those specifically trained to be robust to different robustness benchmark datasets. 
More specifically, we consider PRIME-ResNet50~\cite{PRIME2021} which is trained using new augmentation techniques for enhanced robustness, and FAN-VIT~\cite{zhou2022understanding}, a Vision Transformer with no MLP layers that is highly robust to unseen natural images. 

Additionally, to test transferability for (iii), we use the adversarially generated samples using PRIME-ResNet50 and FAN-VIT as classifiers, and test them on other modern architectures (1) ResNet50~\cite{resnet}, (2) MAE~\cite{MaskedAutoencoders2021}, (3)~RegNet-Y~\cite{regnet2022}, (4)~data2vec~\cite{baevski2022}, (5) ConvNeXt, and (6) ResNeXt.
We then explore a training regime for improving model robustness to our APT manipulations. 
Lastly, for (i-ii), we conduct an ablation study, illustrating the necessity of the different components for generating our samples and investigating the effect of different hyperparameters. 

\subsection{Adversarially Generated Manipulations}

\begin{table}
    \centering
    \resizebox{\linewidth}{!}{  
    \begin{tabular}{lcccccccc}
        \toprule
        Model
        & Real &  & Generated &  \\ 
        \midrule
        PRIME-ResNet50 (APT) & 77.1\% &  & 57.7\% \\
        PRIME-ResNet50 (PGD) & 77.1\% &  & 2.4\% &  \\ 
        PRIME-ResNet50 (SSAH) & 77.1\% &  & 0.7\% & \\ 
        \midrule
FAN-VIT (APT) & 83.6\% &  & 62.0\% &  
        \\
        \bottomrule
    \end{tabular}
    }
    \caption{Accuracy on the ImageNet validation set (Real) and corresponding Generated images with APT, using PRIME-ResNet50~\cite{PRIME2021} in comparison to 
    PGD \cite{madry2017towards} and SSAH \cite{luo2022frequency}. 
    We also consider our method on the FAN-VIT~\cite{zhou2022understanding} classifier. 
    Each model is evaluated on the generated images for which it was also used as the classifier during the generation.
    }
    \vspace*{-3mm}
    \label{tab:results-main}
\end{table}

\begin{table*}
    \vspace{-0.2cm}

    \centering

    \begin{tabular}{llcccccccc}
        \toprule
        \multicolumn{1}{l}{Model} &
        \multicolumn{1}{c}{Real~(Acc)} &
        \multicolumn{1}{c}{Real~(Conf)} &
        \multicolumn{1}{c}{PRIME~(Acc)} &
        \multicolumn{1}{c}{PRIME~(Conf)} &
        \multicolumn{1}{c}{FAN~(Acc)} & 
        \multicolumn{1}{c}{FAN~(Conf)} \\
        \midrule
        PRIME & 77.1\% & 69.7 & 57.7\%$^*$  & 23.4$^*$ & 60.1\% & 52.3 \\
        FAN-VIT & 83.6\% & 62.4 & 70.9\%  & 52.0& 62.0\%$^*$ & 44.7$^*$ \\
        \midrule
        Resnet-50 &  75.3\% & 68.6 & 60.9\%  & 48.3 & 59.7\% & 51.5 \\
        MAE-ViT-H & 87.0\% & 80.1 & 73.8\%  & 65.4 & 62.3\% & 58.4 \\
        Regnet-1280 & 83.7\% & 77.2 & 61.3\%  & 64.5 & 53.0\% & 61.3 \\
        data2vec & 83.5\% & 77.7 & 70.6\%  &  65.0 & 61.6\% & 59.1  \\
        ResNeXt-3 & 82.2\% & 59.9 & 68.6\%  &  48.3 & 61.4\% & 42.2  \\
        ConvNeXt-large & 86.1\% & 78.9 & 72.1\%  &  63.1 & 61.3\% & 53.7 \\        
        VitL16 & 78.7\% & 67.7 & 68.7\%  &  57.2 & - & - \\       
        RegNetX16 & 77.9\% & 73.7 & 64.3\%  &  56.8 & - & - \\
        

        \bottomrule
    \end{tabular}
    \vspace{-0.2cm}

    \caption{Transferability of APT generated samples. For the ImageNet-1k validation set, we consider samples generated to fool a PRIME-Resnet50~\cite{PRIME2021} (PRIME) and a FAN-VIT~\cite{zhou2022understanding} (FAN) pretrained classifier. We then test the accuracy (Acc) and mean softmax probability of the labeled class (Conf) on those samples. The left column indicates the classifier on which we tested the accuracy of real or generated samples. $^*$ indicates the accuracy and confidence of samples generated and tested using the same classifier. 
    }
    \vspace{-0.5cm}
    \label{tab:transfer-study}
\end{table*}

\begin{figure*}
    \centering

    \begin{subfigure}[b]{0.10\textwidth}
         \centering
         \includegraphics[width=\linewidth]{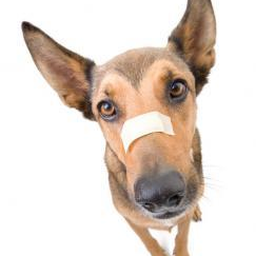}
     \end{subfigure}
     \begin{subfigure}[b]{0.10\textwidth}
         \centering
         \includegraphics[width=\linewidth]{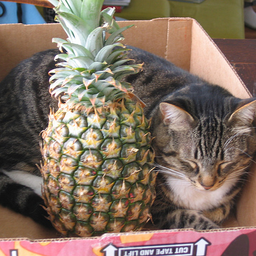}
     \end{subfigure}
     \begin{subfigure}[b]{0.10\textwidth}
         \centering
         \includegraphics[width=\linewidth]{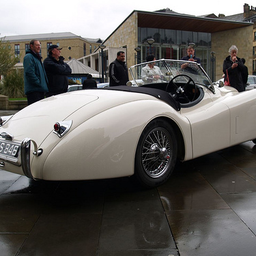}
     \end{subfigure} 
     \begin{subfigure}[b]{0.10\textwidth}
         \centering
         \includegraphics[width=\linewidth]{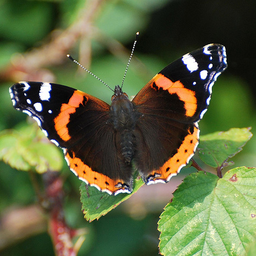}
     \end{subfigure} 
     \begin{subfigure}[b]{0.10\textwidth}
         \centering
         \includegraphics[width=\linewidth]{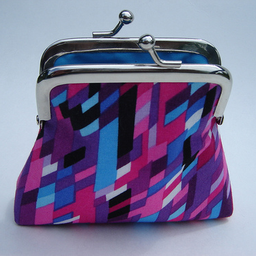}
     \end{subfigure} 
     \begin{subfigure}[b]{0.10\textwidth}
         \centering
         \includegraphics[width=\linewidth]{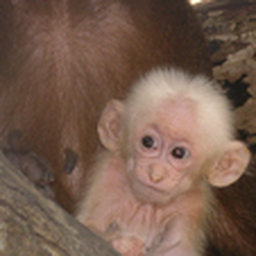}
     \end{subfigure} 
     \begin{subfigure}[b]{0.10\textwidth}
         \centering
         \includegraphics[width=\linewidth]{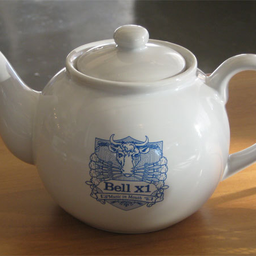}
     \end{subfigure} 
     \begin{subfigure}[b]{0.10\textwidth}
         \centering
         \includegraphics[width=\linewidth]{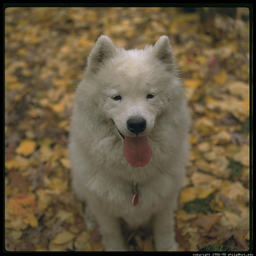}
     \end{subfigure} 
     
    \begin{subfigure}[b]{0.10\textwidth}
         \centering
         \includegraphics[width=\linewidth]{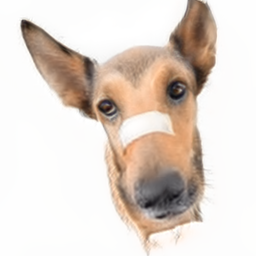}
     \end{subfigure}
     \begin{subfigure}[b]{0.10\textwidth}
         \centering
         \includegraphics[width=\linewidth]{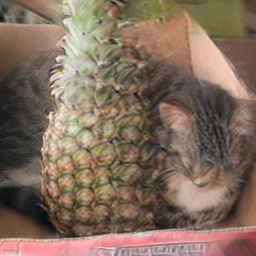}
     \end{subfigure}
     \begin{subfigure}[b]{0.10\textwidth}
         \centering
         \includegraphics[width=\linewidth]{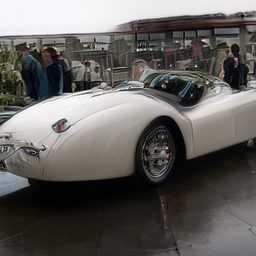}
     \end{subfigure} 
     \begin{subfigure}[b]{0.10\textwidth}
         \centering
         \includegraphics[width=\linewidth]{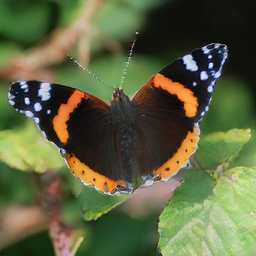}
     \end{subfigure}
     \begin{subfigure}[b]{0.10\textwidth}
         \centering
         \includegraphics[width=\linewidth]{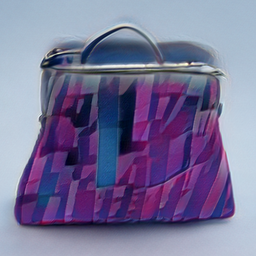}
     \end{subfigure}
     \begin{subfigure}[b]{0.10\textwidth}
         \centering
         \includegraphics[width=\linewidth]{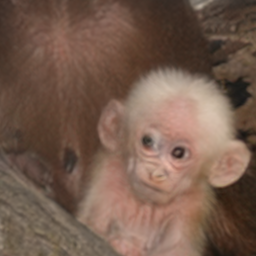}
     \end{subfigure}
     \begin{subfigure}[b]{0.10\textwidth}
         \centering
         \includegraphics[width=\linewidth]{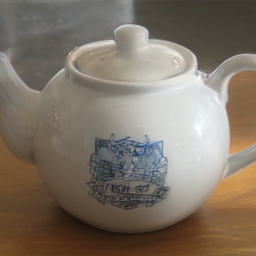}
     \end{subfigure}
     \begin{subfigure}[b]{0.10\textwidth}
         \centering
         \includegraphics[width=\linewidth]{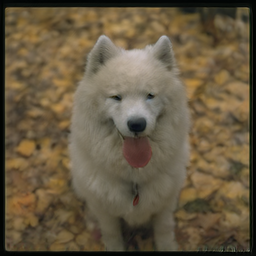}
     \end{subfigure}
     
         \begin{subfigure}[b]{0.10\textwidth}
         \centering
         \includegraphics[width=\linewidth]{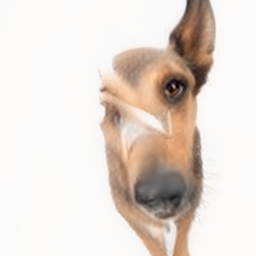}
     \end{subfigure}
     \begin{subfigure}[b]{0.10\textwidth}
         \centering
         \includegraphics[width=\linewidth]{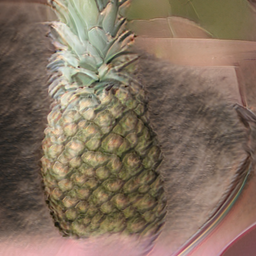}
     \end{subfigure}
     \begin{subfigure}[b]{0.10\textwidth}
         \centering
         \includegraphics[width=\linewidth]{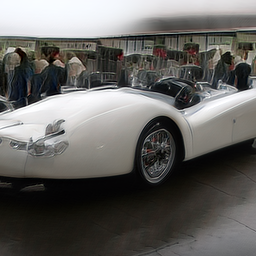}
     \end{subfigure} 
     \begin{subfigure}[b]{0.10\textwidth}
         \centering
         \includegraphics[width=\linewidth]{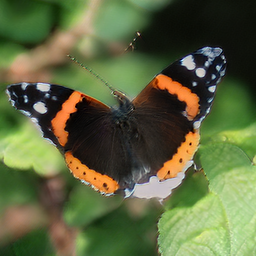}
     \end{subfigure}
     \begin{subfigure}[b]{0.10\textwidth}
         \centering
         \includegraphics[width=\linewidth]{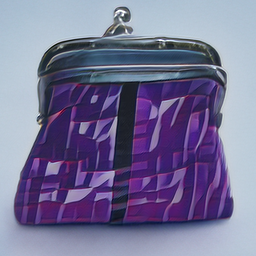}
     \end{subfigure}
     \begin{subfigure}[b]{0.10\textwidth}
         \centering
         \includegraphics[width=\linewidth]{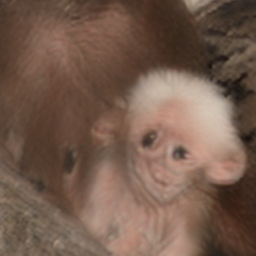}
     \end{subfigure}
     \begin{subfigure}[b]{0.10\textwidth}
         \centering
         \includegraphics[width=\linewidth]{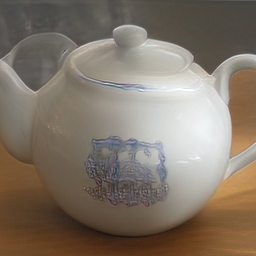}
     \end{subfigure}
     \begin{subfigure}[b]{0.10\textwidth}
         \centering
         \includegraphics[width=\linewidth]{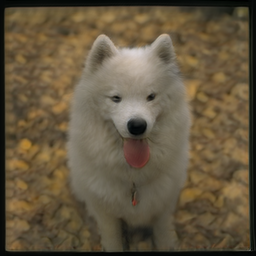}
     \end{subfigure}
    \caption{\textbf{Generated manipulations.}  Top row shows input images. The middle row shows APT manipulations for a ResNet-50 classifier, and the bottom row shows APT manipulations from a FAN-VIT classifier. 
    The leftmost image of a dog and the subsequent images including the image of a butterfly and column 7 (Fluffy dog) show similar manipulations for both classifiers, column 5-6 shows texture and spatial manipulations, the last column showcase a fooling image without a clear APT manipulation.}
    \label{fig:images}


    \centering
    \begin{subfigure}[b]{0.11\textwidth}
         \centering
         \includegraphics[width=\linewidth]{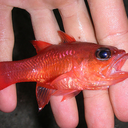}
     \end{subfigure}
     \begin{subfigure}[b]{0.11\textwidth}
         \centering
         \includegraphics[width=\linewidth]{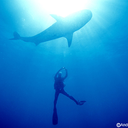}
     \end{subfigure}
     \begin{subfigure}[b]{0.11\textwidth}
         \centering
         \includegraphics[width=\linewidth]{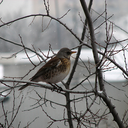}
     \end{subfigure} 
     \begin{subfigure}[b]{0.11\textwidth}
         \centering
         \includegraphics[width=\linewidth]{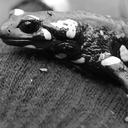}
     \end{subfigure}
    \begin{subfigure}[b]{0.11\textwidth}
         \centering
         \includegraphics[width=\linewidth]{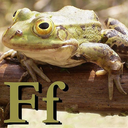}
     \end{subfigure}
     \begin{subfigure}[b]{0.11\textwidth}
         \centering
         \includegraphics[width=\linewidth]{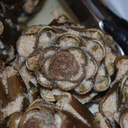}
     \end{subfigure}
     \begin{subfigure}[b]{0.11\textwidth}
         \centering
         \includegraphics[width=\linewidth]{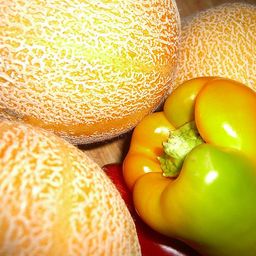}
     \end{subfigure} 

    \begin{subfigure}[b]{0.11\textwidth}
         \centering
         \includegraphics[width=\linewidth]{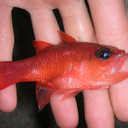}
     \end{subfigure}
     \begin{subfigure}[b]{0.11\textwidth}
         \centering
         \includegraphics[width=\linewidth]{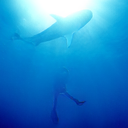}
     \end{subfigure}
     \begin{subfigure}[b]{0.11\textwidth}
         \centering
         \includegraphics[width=\linewidth]{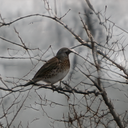}
     \end{subfigure} 
     \begin{subfigure}[b]{0.11\textwidth}
         \centering
         \includegraphics[width=\linewidth]{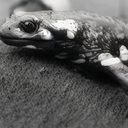}
     \end{subfigure}
    \begin{subfigure}[b]{0.11\textwidth}
         \centering
         \includegraphics[width=\linewidth]{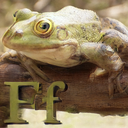}
     \end{subfigure}
     \begin{subfigure}[b]{0.11\textwidth}
         \centering
         \includegraphics[width=\linewidth]{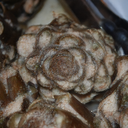}
     \end{subfigure}
     \begin{subfigure}[b]{0.11\textwidth}
         \centering
         \includegraphics[width=\linewidth]{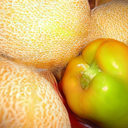}
     \end{subfigure} 

    \begin{subfigure}[b]{0.11\textwidth}
         \centering
         \includegraphics[width=\linewidth]{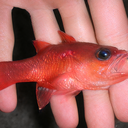}
     \end{subfigure}
     \begin{subfigure}[b]{0.11\textwidth}
         \centering
         \includegraphics[width=\linewidth]{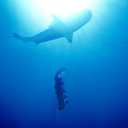}
     \end{subfigure}
     \begin{subfigure}[b]{0.11\textwidth}
         \centering
         \includegraphics[width=\linewidth]{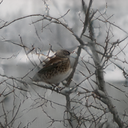}
     \end{subfigure} 
     \begin{subfigure}[b]{0.11\textwidth}
         \centering
         \includegraphics[width=\linewidth]{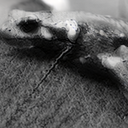}
     \end{subfigure}
    \begin{subfigure}[b]{0.11\textwidth}
         \centering
         \includegraphics[width=\linewidth]{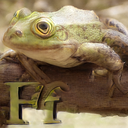}
     \end{subfigure}
     \begin{subfigure}[b]{0.11\textwidth}
         \centering
         \includegraphics[width=\linewidth]{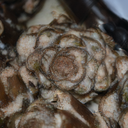}
     \end{subfigure}
     \begin{subfigure}[b]{0.11\textwidth}
         \centering
         \includegraphics[width=\linewidth]{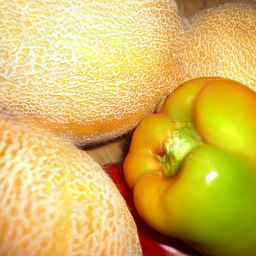}
     \end{subfigure} 

    \begin{subfigure}[b]{0.11\textwidth}
         \centering
         \includegraphics[width=\linewidth]{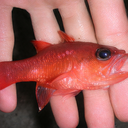}
     \end{subfigure}
     \begin{subfigure}[b]{0.11\textwidth}
         \centering
         \includegraphics[width=\linewidth]{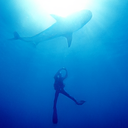}
     \end{subfigure}
     \begin{subfigure}[b]{0.11\textwidth}
         \centering
         \includegraphics[width=\linewidth]{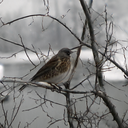}
     \end{subfigure} 
     \begin{subfigure}[b]{0.11\textwidth}
         \centering
         \includegraphics[width=\linewidth]{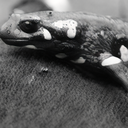}
     \end{subfigure}
    \begin{subfigure}[b]{0.11\textwidth}
         \centering
         \includegraphics[width=\linewidth]{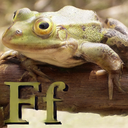}
     \end{subfigure}
     \begin{subfigure}[b]{0.11\textwidth}
         \centering
         \includegraphics[width=\linewidth]{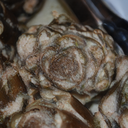}
     \end{subfigure}
     \begin{subfigure}[b]{0.11\textwidth}
         \centering
         \includegraphics[width=\linewidth]{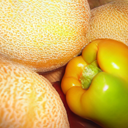}
     \end{subfigure} 
     
    \begin{subfigure}[b]{0.11\textwidth}
         \centering
         \includegraphics[width=\linewidth]{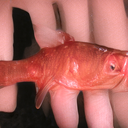}
     \end{subfigure}
     \begin{subfigure}[b]{0.11\textwidth}
         \centering
         \includegraphics[width=\linewidth]{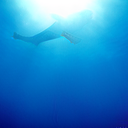}
     \end{subfigure}
     \begin{subfigure}[b]{0.11\textwidth}
         \centering
         \includegraphics[width=\linewidth]{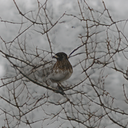}
     \end{subfigure} 
     \begin{subfigure}[b]{0.11\textwidth}
         \centering
         \includegraphics[width=\linewidth]{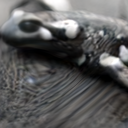}
     \end{subfigure}
    \begin{subfigure}[b]{0.11\textwidth}
         \centering
         \includegraphics[width=\linewidth]{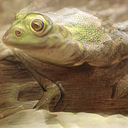}
     \end{subfigure}
     \begin{subfigure}[b]{0.11\textwidth}
         \centering
         \includegraphics[width=\linewidth]{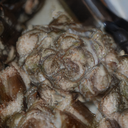}
     \end{subfigure}
     \begin{subfigure}[b]{0.11\textwidth}
         \centering
         \includegraphics[width=\linewidth]{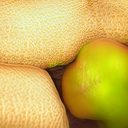}
     \end{subfigure}

    \begin{subfigure}[b]{0.11\textwidth}
         \centering
         \includegraphics[width=\linewidth]{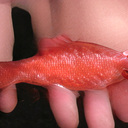}
     \end{subfigure}
     \begin{subfigure}[b]{0.11\textwidth}
         \centering
         \includegraphics[width=\linewidth]{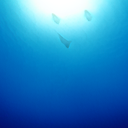}
     \end{subfigure}
     \begin{subfigure}[b]{0.11\textwidth}
         \centering
         \includegraphics[width=\linewidth]{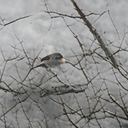}
     \end{subfigure} 
     \begin{subfigure}[b]{0.11\textwidth}
         \centering
         \includegraphics[width=\linewidth]{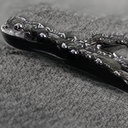}
     \end{subfigure}
    \begin{subfigure}[b]{0.11\textwidth}
         \centering
         \includegraphics[width=\linewidth]{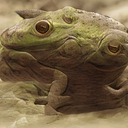}
     \end{subfigure}
     \begin{subfigure}[b]{0.11\textwidth}
         \centering
         \includegraphics[width=\linewidth]{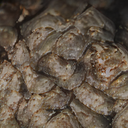}
     \end{subfigure}
     \begin{subfigure}[b]{0.11\textwidth}
         \centering
         \includegraphics[width=\linewidth]{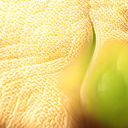}
     \end{subfigure} 
    \caption{\textbf{APT sample generation ablation.} The first row shows the input images. The second row shows APT samples generated using our full objective $\mathcal{L}_{APT}$, all of which fool the pretrained PRIME-ResNet50 classifier. 
    In the third row, we consider $\mathcal{L}_{APT}$ without the reconstruction loss ($\mathcal{L}_{rec}$). In the fourth row, we consider $\mathcal{L}_{APT}$ without the fooling objective ($\mathcal{L}_{CE}$). Without $\mathcal{L}_{CE}$, all but one sample fool the classifier. 
    In the fifth row, we consider $\mathcal{L}_{APT}$ without the discriminator loss ($\mathcal{L}_{PG}$). The sixth row considers $\mathcal{L}_{APT}$ where only the latent space is optimised (generator's parameters fixed), resulting in loss of class preservation.
    \vspace{-0.2cm}
    }
    \vspace{-0.1cm}
    \label{fig:component_examples}

\end{figure*}

\begin{figure}
\includegraphics[width=\linewidth]{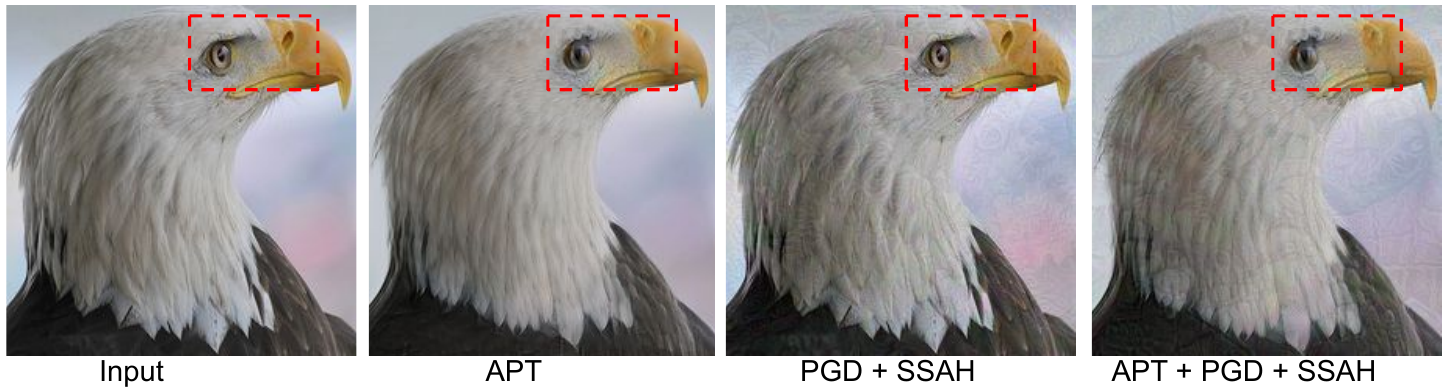}
\caption{Illustration of APT's attack as well as the combination of APT with noise-based attacks (PDF+SSAH). }
\label{fig:new_illus}
\end{figure}

\begin{table}
\small
\centering
\aboverulesep=0.2ex
\belowrulesep=0.2ex
\begin{tabular}{c|c|cc}
    \toprule
    & \textsc{Targeted} & \multicolumn{2}{c}{\textsc{Transfer}} \\
    \midrule
    & PrimeRN50 & VitL16 & RegNetX16 \\
    \midrule
    APT (ours) & 33.7 \% & 16.1 \% & 21.3 \%\\
    \midrule
    SSAH & 80.4 \% & 0.3 \% & 2.4 \%\\
    SSAH + APT (ours) & 95.9 \% & 16.7 \% & 23.8 \%\\
    \midrule
    PGD & 99.6 \% & 13.6 \% & 35.2 \%\\
    PGD + SSAH & 99.8 \% &  13.0 \% & 35.8 \% \\
    PGD + APT (ours) & \textbf{100.0 \%} & \textbf{32.0 \%} & 65.3 \% \\

    \midrule
    ALL & 99.7\% & 31.5 \% & \textbf{65.7\%} \\
    \bottomrule
\end{tabular}

\caption{Attack Success Rate (ASR) for APT, SSAH and PGD. } 
\label{tab:new}

\end{table}

\begin{table}
\centering
\vspace*{-1pt}
\aboverulesep=0.2ex
\belowrulesep=0.2ex
\begin{tabular}{lccc}
    \toprule 
    & APT eval  & PGD eval & SSAH eval \\
    \midrule
    No finetune  & 57.7 \% &  2.4 \% & 0.7 \% \\

    \midrule
    APT finetune  & \textbf{60.2 \%}  & 2.5 \% & \textbf{1.0 \%}\\
    PGD finetune  & 51.2 \%  & \textbf{3.0 \%} & 0.8 \%\\
    SSAH finetune  & 54.6 \%  & 2.3 \% & \textbf{1.0 \%}\\
    \midrule
    All finetune & 56.5 \%  &  \textbf{3.0 \%} & 0.9 \%\\
    \bottomrule
\end{tabular}

\caption{Average accuracy on APT/PGD/SSAH samples using PRIME-ResNet50 before and after fine-tuning on APT/PGD/SSAH samples (or all of them together).}
\label{tab:results-per-severity-imagenet}
\vspace{-0.3cm}
\end{table}


We evaluate our generated samples with respect to  properties (i)-(iv) above. \\

\noindent\textbf{Fidelity and Diversity. \quad}
To measure (i), i.e., whether samples lie on the ImageNet manifold, we are interested in measuring both \textit{Fidelity}--- are the generated samples of high quality?---and \textit{Diversity}--- do the generated samples capture the diversity of the original real dataset?
%
To capture both \textit{Fidelity} and \textit{Diversity} we consider the FID score~\cite{fid}. For a pretrained PRIME-ResNet50 classifier~\cite{PRIME2021}, we consider three groups of images: (1) 3k images chosen at random from the ImageNet validation set, (2) their corresponding adversarial manipulations generated using APT, (3) 3k images chosen at random from the ImageNet training set.

First, we evaluate the FID score between (1) and (2). As can be seen in \cref{tab:FID}, the value is lower than the other groups, indicating that the distributions are close. To evaluate the FID against non-matching groups of real images, we consider the FID between (1) and (3) and between (2) and (3). 
\begin{table}
    \setlength{\tabcolsep}{12pt}
    \centering
    \footnotesize
    \begin{tabular}{cccc}
        \toprule
        & (1) \& (2) 
        & (2) \& (3)  \\ 
        \midrule
        Ours (PRIME) & \textbf{19.87}  & \textbf{23.72}\\
        \cite{lin2020dual}-P (PRIME) & 63.63 & 92.86 \\  
        \cite{lin2020dual}-L (PRIME) & 50.94 & 61.62 \\
        \midrule
        Ours (FAN-VIT) & 20.01  & 24.24 \\
        
        \bottomrule
    \end{tabular}
    \caption{\textbf{Top three rows:}
    FID scores using a PRIME-Resnet50 for our generated manipulations in comparison to manipulations generated by \cite{lin2020dual} using pixel space manipulations (\cite{lin2020dual}-P) on StyleGAN-XL's reconstructions or latent-space manipulations (\cite{lin2020dual}-L). The same set of input images is used. 
    \textbf{Fourth row:} FID scores for our generated samples using a FAN-VIT classifier. The FID score between real validation and training images from ImageNet ((1) \& (3)) is 25.99. }
    \vspace*{-2pt}
    \label{tab:FID}
\end{table}
As can be seen in \cref{tab:FID}, the FID value between (2) and (3) is only slightly higher than that of (1) and (3), indicating that our generated distribution matches the training image distribution in only a slightly worse manner than real validation images.  
We note that the trace of the covariance matrices contributes to the vast majority of the score, likely due to the low number of samples available for the validation set. To verify this we report the FID between the training set and their corresponding adversarial manipulations (43k samples) to be $6.62$ indicating that the real and generated images are similar. 


As a point of comparison, we consider pixel-space adversarial manipulations (\cite{lin2020dual}-P) or latent-space manipulates applied using StyleGAN-XL (\cite{lin2020dual}-L) to generate manipulations (2) on the same set of images. Additionally, we replace StyleGAN-XL with a Diffusion model \cite{nulltextdiff} and as can be seen in \cref{tab:FID} using  \cite{lin2020dual} results in a lower FID score indicating that they are of much lower generation quality and do not match ImageNet's real image distribution. 

\noindent\textbf{Class Preservation. \quad}
To measure (ii), we consider, for a pretrained PRIME-ResNet50 classifier, whether generated samples are class-preserving. We conduct user studies consisting of 25 users and  $40$ samples from ImageNet's validation set and their corresponding samples generated with APT (Q1-3) and \cite{lin2020dual}-P/L (Q3).
We then conduct the following assessments: 
\begin{itemize}
    \item {Q1:} We display each generated sample in isolation and ask how realistic it is, on a Likert scale of 1 (strongly disagree) to 5 (strongly agree).
    \item {Q2:} For each generated sample, we first display the real image and the associated class to the user. We then display the generated sample and ask whether the class is preserved.  
    Additionally, we consider if the pretrained classifier misclassifies the generated image.
    \item {Q3:} We display each image and ask the user to assign a corresponding label. The label is considered correct if it corresponds to the real image's ground truth label. This is performed separately with different users for real images, for our generated samples, and for those generated by \cite{lin2020dual}. 
    
\end{itemize}

For Q1, we report the mean score on the Likert-scale to be $3.55$ 
For Q2, as seen in \cref{tab:fp-rate}, the user almost always states that the class is preserved, except in 5\% of the cases, whereas 32.5\% of these images fool the classifier.
For Q3, as can be seen in \cref{tab:fp-rate}, the generated samples by \cite{lin2020dual}-L/P exhibit 
significant loss of class identity. For APT and the real images, the users correctly predict 90\% and 95\% of the images respectively, and an additional 85\% when using \cite{nulltextdiff} instead of StyleGAN-XL, suggesting that our method yields realistic and class-preserving images.

\begin{table}
    \setlength{\tabcolsep}{10pt}
    \centering
    \footnotesize
    \begin{tabular}{lcc}
        \toprule
         \quad\quad\quad\quad\quad\quad Classifier & (same cl.) & (different cl.)  \\
         \midrule
        Human (same class) & 25 & 13\\
        Human (different class) & 1& 1 \\
        \bottomrule 
    \end{tabular} \\ 
    (Q2) \\ 
    \vspace{0.1cm}
    \begin{tabular}{ccccc}
        \toprule
          \cite{lin2020dual}-L & \cite{lin2020dual}-P & Diffusion &  Ours  
         & Real  \\
         \midrule
         12.5\% & 42.5\% & 85\% & \textbf{90.0\%} & 95.0\% \\
        \bottomrule
    \end{tabular} \\ 
    (Q3) \vspace{-0.1cm}
    \caption{User studies. (Q2). A user study determining if the majority of 25 annotators believe a generated image have changed its class from the real image. (see user study details in \cref{sec:experiments}). Similarly, for the pretrained classifier, we consider if its classification changed. We consider $40$ such samples from the ImageNet-1k validation set extracted using a pretrained PRIME-ResNet50  classifier. (Q3). For our generated samples, real images and those of \cite{lin2020dual}, we display each image and ask the user to assign the corresponding label. The percentage of correct responses corresponding to the real image's class is shown.  }
    \label{tab:fp-rate}
    \vspace{-0.5cm}
\end{table}



\noindent\textbf{Classifier Fooling. \quad}
To test the degree to which a target classifier is fooled, we measure its accuracy on 3k images, 3 for each class, from the ImageNet1k validation set and on the corresponding images generated by APT and other adversarial attacks such as PGD \cite{madry2017towards} and SSAH \cite{luo2022frequency}.
Additionally, we measure the average decrease of the softmax probability for the real class, to assess the decrease in confidence of the classifier on the real class. 
As shown in \cref{tab:results-main}, the accuracy drops by as much as 19.4\%, down to a level comparable to the ImageNet-C accuracy. 
Generating samples using PGD and SSAH 
results in the accuracy dropping by more than 74.7\% and 76.4\% respectvely.
Similarly, in \cref{tab:transfer-study} we assess whether our APT samples are transferable. I.e whether images generated by APT using PRIME-ResNet50 and FAN-VIT classifiers  fool other classifiers.  We observe that the 
fooling images are transferable and the accuracy and confidence on different classifiers also drop.

In Tab.~\ref{tab:new}, we consider the attack success rate (ASR) for APT and noise-baed attacks under the same perturbation magnitude. We consider the ASR for the targeted classifier as well as for a different classifier (transfer).
As our method is complementary to noise-based attacks, using our attacks with noise-based attacks results in the best performance. We provide a visual illustration in \cref{fig:new_illus}.

\noindent\textbf{Diversity of Manipulations. \quad}
In addition to our diverse visual manipulations shown in \cref{fig:teaser}, in \cref{fig:images}, we show for the same set of images, model-dependent APT manipulations which fool either a ResNet50 or a FAN-VIT classifier. Interestingly, for FAN-VIT, other manipulations like texture or spatial transformations (columns 5-6) are more present, in addition to more prevalent versions of the same manipulation as for the ResNet50 classifier (columns 1-4), or no clear manipulation (column 8).

To assess diversity numerically, 
for each image, we consider 5 different APT manipulations by using different seeds. We then consider the 
perceptual distance between each of the generated manipulations (per image) and calculate the mean distance. We average this score over the images. Conceptually, the larger the distance, the larger the diversity. 
For distances d=0.1/0.2/0.4 (resp.), we get a mean of  0.15/0.2/0.4 (resp.).

\subsection{Improving Robustness to Generated Samples}
\label{sec:improving}

We now consider whether APT samples can be used to improve robustness. To this end, for a PRIME-ResNet50 classifier $C$, we use APT, PGD, and SSAH to manipulate 50k images from the ImageNet training set, and fine-tune the classifier on these resulting images (wither in isolation or together), resulting in classifier $C_{finetune}$. 
We then consider 3k images from the ImageNet validation set and generate APT/PGD/SSAH samples for both $C$ and $C_{finetune}$. 
In \cref{tab:results-per-severity-imagenet}, we observe that accuracy on APT-generated images increases by 2.5\% after fine-tuning on APT-generated images. For images generated using PGD and SSAH the accuracy drops, suggesting that these adversarial attacks likely learn a different kind of robustness.
\cref{tab:results-per-severity-imagenet} 
shows that fine-tuning on one set of images results in improved performances on the same type of attack but a reduction in performance compared to other attacks. 
finetuning on all three attacks results in an improved performance.

\begin{figure}
    \centering
    ~\begin{subfigure}[b]{0.10\textwidth}
         \centering
         \includegraphics[width=\linewidth]{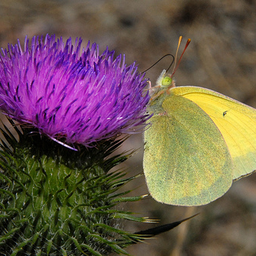}
     \end{subfigure}
     \begin{subfigure}[b]{0.10\textwidth}
         \centering
         ~\includegraphics[width=\linewidth]{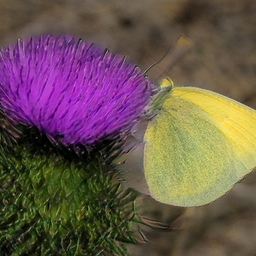}
     \end{subfigure}
     \begin{subfigure}[b]{0.10\textwidth}
         \centering
         ~\includegraphics[width=\linewidth]{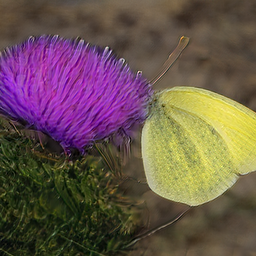}
     \end{subfigure} 
     \begin{subfigure}[b]{0.10\textwidth}
         \centering
         \includegraphics[width=\linewidth]{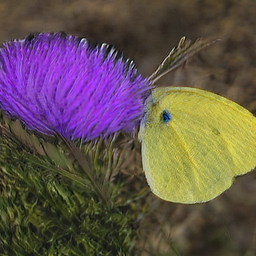} 
     \end{subfigure} \\ 
     \hspace{0.01cm}
    \begin{subfigure}[b]{0.10\textwidth}
         \centering
         \includegraphics[width=\linewidth]{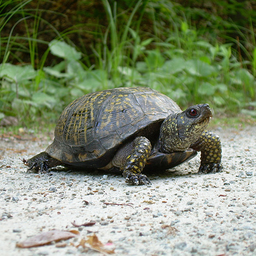}
     \end{subfigure}
     \begin{subfigure}[b]{0.10\textwidth}
         \centering
         \includegraphics[width=\linewidth]{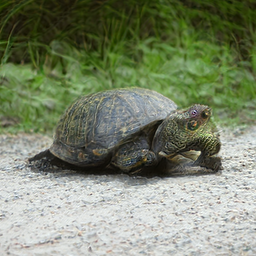}
     \end{subfigure}
     \begin{subfigure}[b]{0.10\textwidth}
         \centering
         \includegraphics[width=\linewidth]{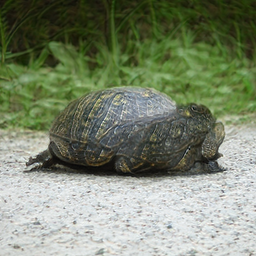}
     \end{subfigure} 
     \begin{subfigure}[b]{0.10\textwidth}
         \centering
         \includegraphics[width=\linewidth]{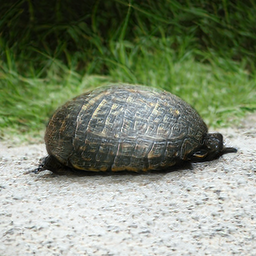}
     \end{subfigure}
    \caption{\textbf{APT generation for various distance $d$ cutoff values}. The leftmost image shows the input image. We increase the maximum distance $d$ to $0.2$, $0.3$ and $0.4$ respectively, for a PRIME-ResNet50 classifier.}
    \label{fig:cutoff_examples}
\end{figure}

\subsection{Ablation Study}

First, we consider the effect of removing each component to our APT objective (\cref{eq:loss}), using a PRIME-ResNet50 classifier.
\cref{fig:component_examples} illustrates examples of generated images with one of the components removed: $\mathcal{L}_{rec}$ (reconstruction), $\mathcal{L}_{PG}$ (discriminator realness) and $\mathcal{L}_{CE}(c_{any};C(G_{\theta}(w_p)))$ (fooling objective). Lastly, we consider applying the optimization of \cref{eq:loss} while modifying the latent space and leaving the generator's parameters fixed. 
When the reconstruction or the discriminator realness components are removed, we observe worse image quality. To measure the effect of each setup we record the number of images that fool the classifier, and observe that when the fooling objective is removed, only one out of the seven shown samples fool the pretrained classifier, whereas all seven samples fool the classifier otherwise.
Corresponding to \cref{fig:component_examples}, we evaluate the differences using cosine similarity of semantic CLIP embeddings (the more similar images are, the more class preserving).
Using the full APT objective we get $91.6\pm 1.6$ similarity (mean and SD). Without $L_{rec}$ (reconstruction loss),  we get $87.1\pm3$. 
Without $L_{CE}$ (fooling objective), we get $92.9\pm1.2$, but note that the resulting images are unable to fool the classifier. Without $L_{PG}$ (discriminator loss), we get $71.0\pm4.7$, and freezing the generator parameters gives $63.9\pm 5.1$. 



\noindent\textbf{Reconstruction vs. fooling trade-off. \quad}
The maximum distance $d$ allowed between the input and generated sample before the optimization is stopped (see \cref{framework}) is an important hyperparameter. 
Increasing this distance may allow for more expressive adversarial manipulations, but this may also result in a change of label for the image. Empirically, we found that $d=0.2$ avoids a change of class. 
We investigate the effect of varying this value in $\{0.2, 0.3, 0.4\}$ and show example generations in \cref{fig:cutoff_examples}. We note that the images tend to lose detail with higher values of $d$, which stems from the fact the reconstruction is poorer. Nonetheless, more diverse manipulations are possible, such as the removal of the antennae on the butterfly.




\section{Conclusion}

We have presented Adversarial Pivotal Tuning, a framework for generating highly expressive adversarial manipulations.
We break with the common assumption that robustness benchmarks are not model specific, or in other words, allow for conducting a new type of robustness study tailored around fooling a particular classifier specifically well. This is achieved by leveraging the full capacity of StyleGAN-XL in generating highly detailed and diverse manipulations. 

We have demonstrated that current robust classifiers, 
are vulnerable to this new type of attack. 
As it turns out, it is possible to also boost performance by using the same framework to create training images as an additional type of augmentation. 
We have shown that APT can successfully be applied both as a way to fool classifiers and as a training framework to improve robustness.
{\small
\bibliographystyle{ieee_fullname}
\bibliography{egbib}

\begin{thebibliography}{10}\itemsep=-1pt

\bibitem{abdal2019image2stylegan}
Rameen Abdal, Yipeng Qin, and Peter Wonka.
\newblock Image2stylegan: How to embed images into the stylegan latent space?
\newblock In {\em Proceedings of the IEEE international conference on computer vision}, pages 4432--4441, 2019.

\bibitem{abdal2020image2stylegan++}
Rameen Abdal, Yipeng Qin, and Peter Wonka.
\newblock Image2stylegan++: How to edit the embedded images?
\newblock In {\em Proceedings of the IEEE/CVF Conference on Computer Vision and Pattern Recognition}, pages 8296--8305, 2020.

\bibitem{akhtar2021advances}
Naveed Akhtar, Ajmal Mian, Navid Kardan, and Mubarak Shah.
\newblock Advances in adversarial attacks and defenses in computer vision: A survey.
\newblock {\em IEEE Access}, 9:155161--155196, 2021.

\bibitem{alaifari2018adef}
Rima Alaifari, Giovanni~S Alberti, and Tandri Gauksson.
\newblock Adef: An iterative algorithm to construct adversarial deformations.
\newblock {\em arXiv preprint arXiv:1804.07729}, 2018.

\bibitem{alcorn2018strike}
Michael~A Alcorn, Qi Li, Zhitao Gong, Chengfei Wang, Long Mai, Wei-Shinn Ku, and Anh Nguyen.
\newblock Strike (with) a pose: Neural networks are easily fooled by strange poses of familiar objects.
\newblock {\em arXiv preprint arXiv:1811.11553}, 2018.

\bibitem{baevski2022}
Alexei Baevski, Wei-Ning Hsu, Qiantong Xu, Arun Babu, Jiatao Gu, and Michael Auli.
\newblock data2vec: A general framework for self-supervised learning in speech, vision and language.
\newblock {\em arXiv preprint arXiv:2202.03555}, 2022.

\bibitem{bhattad2019unrestricted}
Anand Bhattad, Min~Jin Chong, Kaizhao Liang, Bo Li, and David~A Forsyth.
\newblock Unrestricted adversarial examples via semantic manipulation.
\newblock {\em arXiv preprint arXiv:1904.06347}, 2019.

\bibitem{brown2017adversarial}
Tom~B Brown, Dandelion Man{\'e}, Aurko Roy, Mart{\'\i}n Abadi, and Justin Gilmer.
\newblock Adversarial patch.
\newblock {\em arXiv preprint arXiv:1712.09665}, 2017.

\bibitem{creswell2018inverting}
Antonia Creswell and Anil~Anthony Bharath.
\newblock Inverting the generator of a generative adversarial network.
\newblock {\em IEEE transactions on neural networks and learning systems}, 30(7):1967--1974, 2018.

\bibitem{dunn2020evaluating}
Isaac Dunn, Laura Hanu, Hadrien Pouget, Daniel Kroening, and Tom Melham.
\newblock Evaluating robustness to context-sensitive feature perturbations of different granularities.
\newblock {\em arXiv preprint arXiv:2001.11055}, 2020.

\bibitem{engstrom2017rotation}
Logan Engstrom, Brandon Tran, Dimitris Tsipras, Ludwig Schmidt, and Aleksander Madry.
\newblock A rotation and a translation suffice: Fooling cnns with simple transformations.
\newblock {\em arXiv preprint arXiv:1712.02779}, 2017.

\bibitem{fletcher2013practical}
Roger Fletcher.
\newblock {\em Practical methods of optimization}.
\newblock John Wiley \& Sons, 2013.

\bibitem{gowal2020achieving}
Sven Gowal, Chongli Qin, Po-Sen Huang, Taylan Cemgil, Krishnamurthy Dvijotham, Timothy Mann, and Pushmeet Kohli.
\newblock Achieving robustness in the wild via adversarial mixing with disentangled representations.
\newblock In {\em Proceedings of the IEEE/CVF Conference on Computer Vision and Pattern Recognition}, pages 1211--1220, 2020.

\bibitem{regnet2022}
Priya Goyal, Quentin Duval, Isaac Seessel, Mathilde Caron, Ishan Misra, Levent Sagun, Armand Joulin, and Piotr Bojanowski.
\newblock Vision models are more robust and fair when pretrained on uncurated images without supervision, 2022.

\bibitem{guan2020collaborative}
Shanyan Guan, Ying Tai, Bingbing Ni, Feida Zhu, Feiyue Huang, and Xiaokang Yang.
\newblock Collaborative learning for faster stylegan embedding.
\newblock {\em arXiv preprint arXiv:2007.01758}, 2020.

\bibitem{MaskedAutoencoders2021}
Kaiming He, Xinlei Chen, Saining Xie, Yanghao Li, Piotr Doll{\'a}r, and Ross Girshick.
\newblock Masked autoencoders are scalable vision learners.
\newblock {\em arXiv:2111.06377}, 2021.

\bibitem{resnet}
Kaiming He, Xiangyu Zhang, Shaoqing Ren, and Jian Sun.
\newblock Deep residual learning for image recognition.
\newblock {\em CoRR}, abs/1512.03385, 2015.

\bibitem{hendrycks2021natural}
Dan Hendrycks, Kevin Zhao, Steven Basart, Jacob Steinhardt, and Dawn Song.
\newblock Natural adversarial examples.
\newblock In {\em Proceedings of the IEEE/CVF Conference on Computer Vision and Pattern Recognition}, pages 15262--15271, 2021.

\bibitem{fid}
Martin Heusel, Hubert Ramsauer, Thomas Unterthiner, Bernhard Nessler, and Sepp Hochreiter.
\newblock Gans trained by a two time-scale update rule converge to a local nash equilibrium.
\newblock In {\em Proceedings of the 31st International Conference on Neural Information Processing Systems}, NIPS'17, page 6629–6640, Red Hook, NY, USA, 2017. Curran Associates Inc.

\bibitem{hosseini2018semantic}
Hossein Hosseini and Radha Poovendran.
\newblock Semantic adversarial examples.
\newblock In {\em Proceedings of the IEEE Conference on Computer Vision and Pattern Recognition Workshops}, pages 1614--1619, 2018.

\bibitem{joshi2019semantic}
Ameya Joshi, Amitangshu Mukherjee, Soumik Sarkar, and Chinmay Hegde.
\newblock Semantic adversarial attacks: Parametric transformations that fool deep classifiers.
\newblock In {\em Proceedings of the IEEE/CVF international conference on computer vision}, pages 4773--4783, 2019.

\bibitem{karras2020analyzing}
Tero Karras, Samuli Laine, Miika Aittala, Janne Hellsten, Jaakko Lehtinen, and Timo Aila.
\newblock Analyzing and improving the image quality of stylegan.
\newblock In {\em Proceedings of the IEEE/CVF Conference on Computer Vision and Pattern Recognition}, pages 8110--8119, 2020.

\bibitem{laidlaw2020perceptual}
Cassidy Laidlaw, Sahil Singla, and Soheil Feizi.
\newblock Perceptual adversarial robustness: Defense against unseen threat models.
\newblock {\em arXiv preprint arXiv:2006.12655}, 2020.

\bibitem{lee2022autoregressive}
Doyup Lee, Chiheon Kim, Saehoon Kim, Minsu Cho, and Wook-Shin Han.
\newblock Autoregressive image generation using residual quantization.
\newblock In {\em Proceedings of the IEEE/CVF Conference on Computer Vision and Pattern Recognition}, pages 11523--11532, 2022.

\bibitem{lin2020dual}
Wei-An Lin, Chun~Pong Lau, Alexander Levine, Rama Chellappa, and Soheil Feizi.
\newblock Dual manifold adversarial robustness: Defense against lp and non-lp adversarial attacks.
\newblock {\em Advances in Neural Information Processing Systems}, 33:3487--3498, 2020.

\bibitem{lipton2017precise}
Zachary~C Lipton and Subarna Tripathi.
\newblock Precise recovery of latent vectors from generative adversarial networks.
\newblock {\em arXiv preprint arXiv:1702.04782}, 2017.

\bibitem{luo2022frequency}
Cheng Luo, Qinliang Lin, Weicheng Xie, Bizhu Wu, Jinheng Xie, and Linlin Shen.
\newblock Frequency-driven imperceptible adversarial attack on semantic similarity.
\newblock In {\em Proceedings of the IEEE/CVF Conference on Computer Vision and Pattern Recognition}, pages 15315--15324, 2022.

\bibitem{luo2017learning}
Junyu Luo, Yong Xu, Chenwei Tang, and Jiancheng Lv.
\newblock Learning inverse mapping by autoencoder based generative adversarial nets.
\newblock In {\em International Conference on Neural Information Processing}, pages 207--216. Springer, 2017.

\bibitem{madry2017towards}
Aleksander Madry, Aleksandar Makelov, Ludwig Schmidt, Dimitris Tsipras, and Adrian Vladu.
\newblock Towards deep learning models resistant to adversarial attacks.
\newblock {\em arXiv preprint arXiv:1706.06083}, 2017.

\bibitem{PRIME2021}
Apostolos Modas, Rahul Rade, Guillermo {Ortiz-Jim\'enez}, Seyed-Mohsen {Moosavi-Dezfooli}, and Pascal Frossard.
\newblock Prime: A few primitives can boost robustness to common corruptions.
\newblock {\em arXiv preprint arXiv:2112.13547}, 2021.

\bibitem{nulltextdiff}
Ron Mokady, Amir Hertz, Kfir Aberman, Yael Pritch, and Daniel Cohen-Or.
\newblock Null-text inversion for editing real images using guided diffusion models, 2022.

\bibitem{perarnau2016invertible}
Guim Perarnau, Joost Van De~Weijer, Bogdan Raducanu, and Jose~M {\'A}lvarez.
\newblock Invertible conditional gans for image editing.
\newblock {\em arXiv preprint arXiv:1611.06355}, 2016.

\bibitem{poursaeed2021robustness}
Omid Poursaeed, Tianxing Jiang, Harry Yang, Serge Belongie, and Ser-Nam Lim.
\newblock Robustness and generalization via generative adversarial training.
\newblock In {\em Proceedings of the IEEE/CVF International Conference on Computer Vision}, pages 15711--15720, 2021.

\bibitem{qiu2020semanticadv}
Haonan Qiu, Chaowei Xiao, Lei Yang, Xinchen Yan, Honglak Lee, and Bo Li.
\newblock Semanticadv: Generating adversarial examples via attribute-conditioned image editing.
\newblock In {\em European Conference on Computer Vision}, pages 19--37. Springer, 2020.

\bibitem{dalle2}
Aditya Ramesh, Prafulla Dhariwal, Alex Nichol, Casey Chu, and Mark Chen.
\newblock Hierarchical text-conditional image generation with clip latents.
\newblock {\em arXiv preprint arXiv:2204.06125}, 2022.

\bibitem{roich2021pivotal}
Daniel Roich, Ron Mokady, Amit~H Bermano, and Daniel Cohen-Or.
\newblock Pivotal tuning for latent-based editing of real images.
\newblock {\em ACM Trans. Graph.}, 2021.

\bibitem{rombach2021highresolution}
Robin Rombach, Andreas Blattmann, Dominik Lorenz, Patrick Esser, and Björn Ommer.
\newblock High-resolution image synthesis with latent diffusion models, 2021.

\bibitem{Sauer2021NEURIPS}
Axel Sauer, Kashyap Chitta, Jens M{\"{u}}ller, and Andreas Geiger.
\newblock Projected gans converge faster.
\newblock In {\em Advances in Neural Information Processing Systems (NeurIPS)}, 2021.

\bibitem{styleganxl}
Axel Sauer, Katja Schwarz, and Andreas Geiger.
\newblock Stylegan-xl: Scaling stylegan to large diverse datasets, 2022.

\bibitem{shamsabadi2020colorfool}
Ali~Shahin Shamsabadi, Ricardo Sanchez-Matilla, and Andrea Cavallaro.
\newblock Colorfool: Semantic adversarial colorization.
\newblock In {\em Proceedings of the IEEE/CVF Conference on Computer Vision and Pattern Recognition}, pages 1151--1160, 2020.

\bibitem{song2018constructing}
Yang Song, Rui Shu, Nate Kushman, and Stefano Ermon.
\newblock Constructing unrestricted adversarial examples with generative models.
\newblock In {\em Advances in Neural Information Processing Systems}, pages 8312--8323, 2018.

\bibitem{szegedy2013intriguing}
Christian Szegedy, Wojciech Zaremba, Ilya Sutskever, Joan Bruna, Dumitru Erhan, Ian Goodfellow, and Rob Fergus.
\newblock Intriguing properties of neural networks.
\newblock {\em arXiv preprint arXiv:1312.6199}, 2013.

\bibitem{tolosana2020deepfakes}
Ruben Tolosana, Ruben Vera-Rodriguez, Julian Fierrez, Aythami Morales, and Javier Ortega-Garcia.
\newblock Deepfakes and beyond: A survey of face manipulation and fake detection.
\newblock {\em Information Fusion}, 64:131--148, 2020.

\bibitem{xiao2018spatially}
Chaowei Xiao, Jun-Yan Zhu, Bo Li, Warren He, Mingyan Liu, and Dawn Song.
\newblock Spatially transformed adversarial examples.
\newblock In {\em International Conference on Learning Representations}, 2018.

\bibitem{xu2020towards}
Qiuling Xu, Guanhong Tao, Siyuan Cheng, Lin Tan, and Xiangyu Zhang.
\newblock Towards feature space adversarial attack.
\newblock {\em arXiv preprint arXiv:2004.12385}, 2020.

\bibitem{zhang2018unreasonable}
Richard Zhang, Phillip Isola, Alexei~A. Efros, Eli Shechtman, and Oliver Wang.
\newblock The unreasonable effectiveness of deep features as a perceptual metric, 2018.

\bibitem{zhou2022understanding}
Daquan Zhou, Zhiding Yu, Enze Xie, Chaowei Xiao, Anima Anandkumar, Jiashi Feng, and Jose~M. Alvarez.
\newblock Understanding the robustness in vision transformers.
\newblock In {\em International Conference on Machine Learning (ICML)}, 2022.

\end{thebibliography}
}

\end{document}